\documentclass{article} 
\usepackage{iclr2024_conference_arxiv,times}

\usepackage[pagebackref=true,breaklinks=true,letterpaper=true,colorlinks,bookmarks=false]{hyperref}

\usepackage{url}

\usepackage[utf8]{inputenc} 
\usepackage[T1]{fontenc}    
\usepackage{graphicx}
\usepackage{url}            
\usepackage{booktabs}       
\usepackage{amsfonts}       
\usepackage{nicefrac}       
\usepackage{microtype}      
\usepackage{xcolor}         
\usepackage{multicol}
\usepackage{multirow}
\usepackage{pifont}
\usepackage{caption}
\usepackage{amsmath}
\usepackage{listings}
\usepackage[inline]{enumitem}
\usepackage{algorithm}
\usepackage{color, colortbl}
\usepackage{placeins}
\usepackage{caption}

\usepackage{url}
\usepackage{booktabs}       
\usepackage{nicefrac}       
\usepackage{microtype}      
\usepackage{xcolor}         
\usepackage{epsfig}
\usepackage{times}
\usepackage{rotating}
\usepackage{multirow}
\usepackage{tabulary}
\usepackage{float}
\usepackage{textcomp}
\usepackage{subfig}
\usepackage{verbatim}
\usepackage{wrapfig}

\definecolor{cyberpink}{RGB}{180,100,182}

\usepackage{hyperref}
 \hypersetup{
     allcolors = cyberpink
     }

\definecolor{lime}{RGB}{205, 237, 250}

\definecolor{pink}{RGB}{245,220,250}
\definecolor{lemon}{RGB}{255, 255, 171}

\newcommand{\ours}{SlowFormer (ours) }
\newcommand{\sfo}{SlowFormer }
\newcommand{\avit}{A-ViT }
\newcommand{\ignore}[1]{}
\newcommand{\trm}[1]{\textrm{#1}}

\DeclareMathOperator{\softmax}{Softmax}

\usepackage[capitalize]{cleveref}
\crefname{section}{Sec.}{Secs.}
\Crefname{section}{Section}{Sections}
\Crefname{table}{Table}{Tables}
\crefname{table}{Tab.}{Tabs.}

\newcommand\blfootnote[1]{%
  \begingroup
  \renewcommand\thefootnote{}\footnote{#1}%
  \addtocounter{footnote}{-1}%
  \endgroup
}

\usepackage{url}

\title{SlowFormer: Universal Adversarial Patch for Attack on Compute and Energy Efficiency of Inference Efficient Vision Transformers}
\author{
\and
\hspace{-0.28in}
K L Navaneet $^{*}$ \quad
Soroush Abbasi Koohpayegani $^{*}$ \quad
Essam Sleiman $^{*}$ \quad
Hamed Pirsiavash 
\\
University of California, Davis\\
}

\iclrfinalcopy 
\begin{document}

\maketitle

\begin{abstract}
Recently, there has been a lot of progress in reducing the computation of deep models at inference time. These methods can reduce both the computational needs and power usage of deep models. Some of these approaches adaptively scale the compute based on the input instance. We show that such models can be vulnerable to a universal adversarial patch attack, where the attacker optimizes for a patch that when pasted on any image, can increase the compute and power consumption of the model. We run experiments with three different efficient vision transformer methods showing that in some cases, the attacker can increase the computation to the maximum possible level by simply pasting a patch that occupies only 8\% of the image area. We also show that a standard adversarial training defense method can reduce some of the attack's success. We believe adaptive efficient methods will be necessary for the future to lower the power usage of deep models, so we hope our paper encourages the community to study the robustness of these methods and develop better defense methods for the proposed attack. Our code is available here: \textcolor{magenta}{\href{https://github.com/UCDvision/SlowFormer}{https://github.com/UCDvision/SlowFormer}} 
\end{abstract}

\section{Introduction}
\blfootnote{* equal contribution}

The field of deep learning has recently made significant progress in improving the efficiency of inference time. Two broad categories of methods can be distinguished: 1) those that reduce computation regardless of input, and 2) those that reduce the computation depending on the input (adaptively). Most methods, such as weight pruning or model quantization, belong to the first category, which reduces computation by a constant factor, regardless of the input. However, in many applications, the complexity of the perception task may differ depending on the input. For example, when a self-driving car is driving between lanes in an empty street, the perception may be simpler and require less computation when compared to driving in a busy city street scene. Interestingly, in some applications, simple scenes such as highway driving may account for the majority of the time. Therefore, we believe that adaptive computation reduction will become an increasingly important research area in the future, especially when non-adaptive methods reach the lower bound of computation.

We note that reducing computation has at least two advantages: reducing the run time and also reducing the power consumption. We acknowledge that depending on the hardware architecture, reducing the run-time for some input images may not be highly valuable since the system parameters (e.g., camera frame rate) should be designed for the worst-case scenario. Additionally, it might not be possible for other processes to effectively utilize the freed compute cores. However, we argue that reduction of compute usually reduces power usage, which is crucial, particularly in mobile devices that run on battery. This becomes even more important as battery storage technology is not growing as fast as compute technology. For instance, increasing the size of the battery for a drone may lead to a dramatic reduction in its range due to the increased battery weight. 

Assuming that a perception method is reducing the computation adaptively with the input, an adversary can trick the model by modifying the input to increase the computation and power consumption. We are interested in designing a universal adversarial patch that when pasted on any input image, will increase the computation of the model leading to increased power consumption. We believe this is an important vulnerability, particularly for safety-critical mobile systems that run on battery.

As an example, a delivery robot like Starship uses a 1,200Wh battery and can run for 12 hours \citep{starship}, so it uses almost 100 watts for compute and mobility. Hence, an adversary increasing the power consumption of the perception unit by 20 watts, will reduce the battery life by almost 20\%, which can be significant. Note that 20 watts increase in power is realistic assuming that it uses two NVIDIA Jetson Xavier NX cards (almost 20 watts each) to handle its 12 cameras and other sensors.

\begin{figure*}[t!]
    \centering
    
        \centering
        \includegraphics[width=1.0\linewidth]{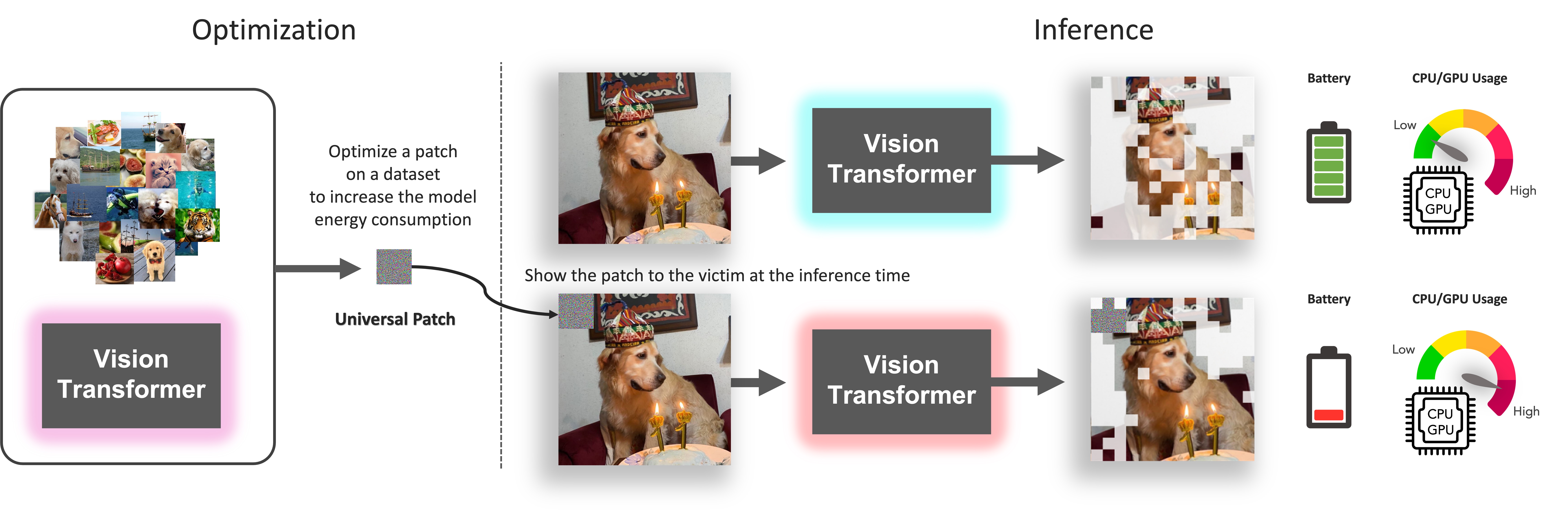}
    
    \caption{\textbf{Energy Attack on Vision Transformers:} Given a pre-trained input-dependent computation efficient model, the adversary first attaches an adversarial patch to all images in a dataset and optimizes this patch with our method such that it maximizes the model's computation for each sample. During inference, the adversary modifies the input of the victim's model by applying the learnt patch to it. This results in an increase in compute in the victim's model. The attack will thus potentially slowdown and also lead to increased energy consumption and CPU/GPU usage on the victim's device.}
    \label{fig:teaser}
\end{figure*}

Please note that in this paper, we do not experiment with real hardware to measure the power consumption. Instead, we report the change in FLOPs of the inference time assuming that the power consumption is proportional to the number of FLOPs.

We design our attack, SlowFormer, for three different methods (A-VIT \citep{yin2022vit}, ATS \citep{fayyaz2022adaptive}, and Ada-VIT \citep{meng2022adavit}) that reduce the computation of vision transformers. These methods generally identify the importance of each token for the final task and drop the insignificant ones to reduce the computation. We show that in all three cases, our attack can increase the computation by a large margin, returning it to the full-compute level (non-efficient baseline) for all images in some settings. Figure \ref{fig:teaser} shows our attack.

There are some prior works that design a pixel-level perturbation attack to increase the compute of the model. However, we believe universal patch-based attacks that do not change with the input image (generalize from training data to test data) are much more practical in real applications. Note that to modify the pixel values on a real robot, the attacker needs to access and manipulate the image between the camera and compute modules, which is impossible in many applications.

{\bf Contributions:} We show that efficient vision transformer methods are vulnerable to a universal patch attack that can increase their compute and power usage. We demonstrate this through experiments on three different efficient transformer methods. We show that an adversarial training defense can reduce attack success to some extent. 

\section{Related Work}

\textbf{Vision Transformers:} The popularity of transformer \citep{vaswani2017attention} architecture in vision has grown rapidly since the introduction of the first vision transformer \citep{dosovitskiy2020image,touvron2021deit}. Recent works demonstrate the strength of vision transformers on a variety of computer vision tasks \citep{vit, deit, swin, deepvit, global_filter_net, detr, setr, maskformer, pointr, pointtransformer}. Moreover, transformers are the backbone of recent Self-Supervised Learning (SSL) models \citep{he2021masked,caron2021emerging}, and vision-language models \citep{clip@clip}. In our work, we design an attack to target the energy and computation efficiency of vision transformers.  

\textbf{Efficient Vision Transformers:} Due to the recent importance and popularity of vision transformers, many works have started to study the efficiency of vision transformers \citep{yu2022metaformer, brown2022dartformer,keles2022computational}. To accomplish this, some lines of work study token pruning with the goal of removing uninformative tokens in each layer \citep{fayyaz2022adaptive,rao2021dynamicvit,marin2021token,yin2022vit,meng2022adavit}. ToMe~\citep{bolya2022token} merges similar tokens in each layer to decrease the computation. Some works address quadratic computation of self-attention module by introducing linear attention \citep{lu2021soft,katharopoulos2020transformers,shen2021efficient,ali2021xcit,koohpayegani2022sima}. Efficient architectures \citep{liu2021swin,ho2019axial} that limit the attention span of each token have been proposed to improve efficiency. 
In our paper, we attack token pruning based efficient transformers where the computation varies based on the input samples~\citep{meng2022adavit,fayyaz2022adaptive,yin2022vit}. 

\textbf{Dynamic Computation:}
There are different approaches to reducing the computation of vision models, including knowledge distillation to lighter network \citep{hinton2015distilling,lu2020twinbert}, model quantization \citep{rastegari2016xnornet,liu2022ecoformer} and model pruning \citep{li2016pruning}. In  these methods, the computation is fixed during inference. In contrast to the above models, some works address efficiency by having variable computation based on the input. The intuition behind this direction is that not all samples require the same amount of computation. 
Several recent works have developed models that dynamically exit early or skip layers \citep{huang2017multi, teerapittayanon2016branchynet, bolukbasi2017adaptive, graves2016adaptive,  wang2018skipnet, veit2018convolutional, guan2017energy, elbayad2019depth, figurnov2017spatially} and selectively activate neurons, channels or branches for dynamic width \citep{cai2021dynamic, fedus2021switch, yuan2020s2dnas, hua2019channel, gao2018dynamic, herrmann2020channel, bejnordi2019batch, chen2019self} depending on the complexity of the input sample. Zhou et al. show that not all locations in an image contribute equally to the predictions of a CNN model \citep{zhou2016learning}, encouraging a new line of work to make CNNs more efficient through spatially dynamic computation. Pixel-Wise dynamic architectures \citep{ren2018sbnet, fan2019blvnet, kong2019pixel, cao2019seernet, verelst2020dynamic, xie2020spatially, chen2021dynamic} learn to focus on the significant pixels for the required task while Region-Level dynamic architectures perform adaptive inference on the regions or patches of the input \citep{li2017dynamic, fu2017look}. Finally, lowering the resolution of inputs decreases computation, but at the cost of performance. Conventional CNNs process all regions of an image equally, however, this can be inefficient if some regions are ``easier" to process than others \citep{howard2017mobilenets}. Correspondingly, \citep{yang2020resolution, yang2019dynamic} develop methods to adaptively scale the resolution of images.

Transformers have recently become extremely popular for vision tasks, resulting in the release of a few input-dynamic transformer architectures \citep{yin2022vit, ats, meng2022adavit}. Fayyaz et al. \citep{ats} introduce a differentiable parameter-free Adaptive Token Sampler (ATS) module which scores and adaptively samples significant tokens. ATS can be plugged into any existing vision transformer architecture. A-VIT \citep{yin2022vit} reduces the number of tokens in vision transformers by discarding redundant spatial tokens. Meng et al. \citep{meng2022adavit} propose AdaViT, which trains a decision network to dynamically choose which patch, head, and block to keep/activate throughout the backbone. 

\textbf{Adversarial Attack:}
Adversarial attacks are designed to fool models by applying a targeted perturbation or patch on an image sample during inference \citep{szegedy2013intriguing,goodfellow2014explaining, kurakin2018adversarial}. These methods can be incorporated into the training set and optimized to fool the model. Correspondingly, defenses have been proposed to mitigate the effects of these attacks \citep{papernot2016distillation, xie2017adversarial, feinman2017detecting, li2017adversarial}. Patch-Fool~\citep{fu2022patch} considers adversarial patch-based attacks on transformers. Most prior adversarial attacks target model accuracy, ignoring model efficiency.

\textbf{Energy Attack:} Very recently, there have been a few works on energy adversarial attacks on neural networks. In ILFO \citep{haque2020ilfo}, Haque et al. attack two CNN-based input-dynamic methods: SkipNet \citep{wang2018skipnet} and SACT \citep{figurnov2017spatially} using image specific perturbation.
DeepSloth \citep{hong2020panda} attack focuses on slowing down early-exit methods, reducing their energy efficiency by 90-100\%. 
GradAuto \citep{pan2022gradauto} successfully attacks methods that are both dynamic width and dynamic depth.
NICGSlowDown and TransSlowDown \citep{chen2022nicgslowdown, chen2021transslowdown} attack neural image caption generation and neural machine translation methods, respectively. All these methods primarily employ image specific perturbation based adversarial attack. SlothBomb injects efficiency backdoors to input-adaptive dynamic neural networks \citep{chenslothbomb} and NodeAttack \citep{haque2021nodeattack} attacks Neural Ordinary Differential Equation models, which use ordinary differential equation solving to dynamically predict the output of a neural network. 
Our work is closely related to ILFO \citep{haque2020ilfo}, DeepSloth \citep{hong2020panda} and GradAuto \citep{pan2022gradauto} in that we attack the computational efficiency of networks. However, unlike these methods, we focus on designing an adversarial patch-based attack that is universal and on vision transformers. We additionally provide a potential defense for our attack. We use a patch that generalizes from train to test set and thus we do not optimize per sample during inference. Our patch-based attack is especially suited for transformer architectures~\citep{fu2022patch}.

\vspace{-.1in}
\section{Energy Attack}

\subsection{Threat Model:} We consider a scenario where the adversary has access to the victim's trained deep model and modifies its input such that the energy consumption of the model is increased. To make the setting more practical, instead of perturbing the entire image, we assume that the adversary can modify the input image by only pasting a patch \citep{brown2017adversarial,Saha2019AdversarialPE} on it and that the patch is universal, that is, image independent. During inference, a pretrained patch is pasted on the test image before propagating it through the network.

In this paper, we attack three state-of-the-art efficient transformers. Since the attacker manipulates only the input image and not the network parameters, the attacked model must have dynamic computation that depends on the input image. As stated earlier, several recent works have developed such adaptive efficient models and we believe that they will be more popular in the future due to the limits of non-adaptive efficiency improvement.

\subsection{Attack on Efficient Vision Transformers:}
\noindent \textbf{Universal Adversarial Patch:} We use an adversarial patch to attack the computational efficiency of transforms. The learned patch is universal, that is, a single patch is
trained and is used during inference on all test images. The patch optimization is performed only on the train set and there is no per-sample optimization on the test images. The patch is pasted on an image by replacing the image pixels using the patch. We assume the patch location does not change from train to test. The patch pixels are 
initialized using IID samples from a uniform distribution over $[0, 255]$. During each training 
iteration, the patch is pasted on the mini-batch samples and is updated to increase the computation of the attacked network. The patch values are projected onto $[0, 255]$ and quantized using $256$ uniform levels after each iteration. Note that the parameters of the network being attacked are not updated during patch training. During inference, the trained patch is pasted on the test images and the computational efficiency of the network on the adversarial image is measured. Below, we describe in detail the efficient methods under attack and the loss formulation used to update the patch. 

Here, we focus on three methods employing vision transformers for the task of image classification. 
All these methods modify the computational flow of the network based on the input image for faster inference. 
A pretrained model is used for the attack and is not modified during our adversarial patch training.
For clarity, we first provide a brief background of each method before describing our attack.\\

\noindent \textbf{Attacking \avit:}

{\bf Background:} \avit~\citep{yin2022vit} adaptively prunes image tokens to achieve speed-up in inference with minimal 
loss in accuracy. For a given image, a dropped token will not be used again in the succeeding layers of 
the network. 
Let $x$ be the input 
image and $\{{t}^l\}_{1:K}$ be the corresponding $K$ tokens at layer $l$. An input-dependent halting score
$h_k^l$ for a token $k$ at layer $l$ is calculated and the token is dropped at layer $N_k$ where its
cumulative halting score exceeds a fixed threshold value $1 - \epsilon$ for the first time. The token is propagated
until the final layer if its score never exceeds the threshold. Instead of introducing a new activation for $h_k^l$, the first dimension of each token is used
to predict the halting score for the corresponding token. The network is trained to maximize the 
cumulative halting score at each layer and thus drop the tokens earlier.
The loss termed ponder loss, is given by:
\begin{equation}
    \begin{split}
        \mathcal{L}_{\textrm{ponder}} = \frac{1}{K}\sum_{k=1}^K (N_k + r_k), \quad \quad
        r_k = 1 - \sum_{l=1}^{N_{k-1}} h_k^l
    \end{split}
    \label{eqn:loss_avit}
\end{equation}
Additionally, \avit enforces a Gaussian prior on the expected halting scores of all tokens via 
$KL$-divergence based distribution loss, $\mathcal{L}_{\trm{distr.}}$. These loss terms are minimized 
along with the task-specific loss $\mathcal{L}_{\trm{task}}$. Thus, the overall training objective is 
$\mathcal{L} = \mathcal{L}_{\trm{task}} + \alpha_d \mathcal{L}_{\trm{distr.}} + 
\alpha_p \mathcal{L}_{\trm{ponder}}$ where $\alpha_d$ and $\alpha_p$ are hyperparameters. 

{\bf Attack:} Here, we train the patch to increase the inference compute of a trained \avit model. Since we are 
interested in the compute and not task-specific performance, we simply use 
$-(\alpha_d \mathcal{L}_{\trm{distr.}} +  \alpha_p \mathcal{L}_{\trm{ponder}})$ as our loss.
It is possible to preserve (or hurt) the task performance by additionally using
$+ \mathcal{L}_{\trm{task}}$ (or $- \mathcal{L}_{\trm{task}}$) in the loss formulation. \\

\noindent \textbf{Attacking AdaViT:} 

\textbf{Background:} To improve the inference efficiency of vision transformers, AdaViT \citep{meng2022adavit} inserts and trains a decision network before each transformer block to dynamically decide which patches, self-attention heads, and transformer blocks to keep/activate throughout the backbone. The $l^{\trm{th}}$ block's decision network consists of three linear layers with parameters ${W}_l$ = ${W}_l^p, {W}_l^h, {W}_l^b$ which are then multiplied by each block's input  ${Z}_l$ to get $m$.

\begin{equation}
    \label{eq:AdaViTWeights}
    {({m}_l^p, {m}_l^h, {m}_l^b)} = {({W}_l^p, {W}_l^h, {W}_l^b) {Z}_l}
\end{equation}

The value $m$ is then passed to sigmoid function to convert it to a probability value used to make the binary decision of keep/discard. Gumbel-Softmax trick \citep{maddison2016concrete}
is used to make this decision differentiable during training. Let $M$ be the keep/discard mask after applying Gumbel-Softmax on $m$. The loss on computation is given by:
\begin{equation}
    \label{eq:AdaViTUsageLoss}
    \begin{split}
        \mathcal{L}_{usage} &= (\frac{1}{{D}_p} \sum_{d=1}^{{D}_p} {M}_d^p - \gamma_p)^2 + (\frac{1}{{D}_h} \sum_{d=1}^{{D}_h} {M}_d^h - \gamma_h)^2 
        + (\frac{1}{{D}_b} \sum_{d=1}^{{D}_b} {M}_d^b - \gamma_b)^2
    \end{split}
\end{equation}

\noindent where $D_p$, $D_h$, $D_b$ represent the number of total patches, heads, and blocks of the entire transformer, respectively. $\gamma_p$, $\gamma_h$, $\gamma_b$ denote the target computation budgets i.e. the percentage of  patches/heads/blocks to keep. The total loss is a combination of task loss (cross-entropy) and computation loss: 
$L = \mathcal{L}_{ce} + \mathcal{L}_{usage}$.

\textbf{Attack:} To attack this model, we train the patch to maximize the computation loss $\mathcal{L}_{usage}$. More specifically, we set the computation-target $\gamma$ values to 0 and negate the $\mathcal{L}_{usage}$ term in Eq.~\ref{eq:AdaViTUsageLoss}. As a result, the patch is optimized to maximize the probability of keeping the corresponding patch (p), attention head (h), and transformer block (b). We can also choose to attack the prediction performance by selectively including or excluding the $\mathcal{L}_{ce}$ term. Note that the computation increase for this method is not as high as for the other methods. To investigate, we ran a further experiment using a patch size of 224x224 (entire image size) to find the maximum possible computation. This resulted in 4.18 GFLOPs on the ImageNet-1K validation set, which is lower than 4.6. If we use this as an upper-bound of GFLOPs increase, our method instead achieves a 49\% Attack Success. \\

\noindent \textbf{Attacking ATS:} 

\textbf{Background:} 
Given $N$ tokens with the first one as the classification token, the transformer attention matrix $\mathcal{A}$ is calculated by the following dot product: 
$\mathcal{A} = \softmax\left(\mathcal{QK}^T / \sqrt{d}\right)$
where $\sqrt{d}$ is a scaling coefficient, $d$ is the dimension of tokens, $\mathcal{Q}$, $\mathcal{K}$ and  $\mathcal{V}$ are the query, key and value matrices, respectively. The value $\mathcal{A}_{1,j}$ denotes the attention of the classification token to token $j$. ATS~\citep{fayyaz2022adaptive} assigns importance score $S_j$ for each token $j$ by measuring how much the classification token attends to it:
\begin{equation}
    \label{eq:score}
    \mathcal{S}_j = \frac{\mathcal{A}_{1,j}\times ||\mathcal{V}_{j}||}{\sum_{i=2} \mathcal{A}_{1,i}\times ||\mathcal{V}_{i}||}
\end{equation}
The importance scores are converted to probabilities and are used to sample tokens, where tokens with a lower score have more of a chance of being dropped. 

\textbf{Attack:} Since ATS uses inverse transform sampling, it results in fewer samples if the importance distribution is sharp. 
To maximize the computation in ATS, we aim to obtain a distribution of scores with high entropy to maximize the number of retained tokens.
Therefore, we optimize the patch so that the attention of the classification token over other tokens is a uniform distribution using the following MSE loss: 
\vspace{-.1in}
$$\mathcal{L}=\sum_{i=2}^{N} ||\mathcal{A}_{1,i}-\frac{1}{N}||_2^2$$

Note that one can optimize $\mathcal{S}$ to be uniform, but we found the above loss to be easier to optimize. For a multi-head attention layer, we calculate the loss for each head and then sum the loss over all heads. Moreover, ATS can be applied to any layer of vision transformers. For a given model, we apply our loss at all ATS layers and use a weighted summation for optimization.

\vspace{-.1in}
\section{Defense}
\label{sec:defense}
We adopt standard adversarial training as a defense method for our attack. In the standard way, at each iteration of training the model, one would load an image, attack it, and then use it with correct labels in training the model. We cannot adopt this out-of-the-box since our attack generalizes across images and is not dependent on a single image only. To do this, we maintain a set of adversarial patches, and at each iteration sample one of them randomly (uniformly), and use it at the input while optimizing the original loss of the efficient model to train a robust model. To adapt the set of adversarial patches to the model being trained, we interrupt the training at every 20\% mark of each epoch and optimize for a new patch to be added to the set of patches. To limit the computational cost of training, we use only 500 iterations to optimize for a new patch, which results in an attack with reasonable accuracy compared to our main results.

\vspace{-.1in}
\section{Experiments}

\subsection{Attack on Efficient Vision Transformers}
\noindent \textbf{Dataset:} We  evaluate the effectiveness of our attack on two datasets: ImageNet-1K \citep{deng2009imagenet} and CIFAR-10 \citep{cifar}. ImageNet-1K contains $1.3$M images in the train set and $50$K images in the validation set with $1000$ total categories. CIFAR-10 has $50$K images for training and $10$K images for validation with $10$ total categories. \\

\noindent \textbf{Metrics:} We report Top-1 accuracy and average computation in terms of GFLOPs for both attacked and unattacked models. Similar to Attack Success Rate in a standard adversarial attack, we introduce a metric: Attack Success to quantify the efficacy of the attack. We define Attack Success as the number of FLOPs increased by the attack divided by the number of FLOPs decreased by the efficient method. $\trm{Attack Success} = \frac{(\trm{FLOPs}_{\trm{attack}} - \trm{FLOPs}_{\trm{min}})} {(\trm{FLOPs}_{\trm{max}} - \trm{FLOPs}_{\trm{min}})}$ where $\trm{FLOPs}_{\trm{min}}$ is the compute of the efficient model and $\trm{FLOPs}_{\trm{max}}$ is that of the original inefficient model. Attack Success is thus capped at $100\%$ while a negative value denotes a reduction in FLOPs. Note that our Attack Success metric illustrates the effectiveness of an attack in reversing the FLOPs reduction of a particular method.\\

\begin{table}[!htb]
    \begin{minipage}[t]{0.55\linewidth}
        \caption{\textbf{Energy Attack on Efficient Vision Transformers:} Comparison of the effect of energy attack with baselines: No Attack, Random Patch, targeted (TAP), and non-targeted (NTAP) adversarial patches applied to three input-dynamic computation efficient pre-trained models of varying architectures. The maximum possible compute for a given architecture is provided in bold. On \avit, we completely undo the efficiency gains obtained by the efficient method through our attack, achieving Attack Success of $100\%$. We achieve high Attack Success on all approaches while the baselines expectedly do not contribute to increase in compute.
        }
        \label{tab:imagenet_attack}
        \centering
        \scalebox{0.83}
        {
        \begin{tabular}{lcccc} %
        \toprule
        Method & Attack &  Model &  Top-$1$ & Attack \\
         & &  GFLOPs & Acc & Success \\
        \midrule 
        \midrule
                                &  \textbf{ViT-Tiny}            & \textbf{1.3}  & - & -\\
                                \cline{2-5}
        \multirow{5}{*}{A-ViT} & No attack    & 0.87 & 71.4\% & -\\
                                & Random Patch  & 0.87 & 70.8\% & -1\%\\
                                & TAP           & 0.85 & 0.1\%  & -5\%\\
                                & NTAP          & 0.83 & 0.1\%  & -10\%\\
                                \rowcolor{pink}
                                & \ours         & 1.3 & 4.7\% & 100\%\\
        \midrule
                                &  \textbf{ViT-Small}            & \textbf{4.6}  & - & -\\
                                \cline{2-5}
        \multirow{5}{*}{A-ViT} & No attack    & 3.7    & 78.8\% & - \\
                                & Random Patch  & 3.7  & 78.4\% & -2\%\\
                                & TAP           & 3.6  & 0.1\%  & -12\%   \\
                                & NTAP          & 3.6  & 0.1\%  & -7\%\\
                                \rowcolor{pink}
                                & \ours         & 4.6  & 2.3\%  & 99\%\\
        \midrule
                                &  \textbf{ViT-Tiny}            & \textbf{1.3}  & -  & - \\
                                \cline{2-5}
        \multirow{5}{*}{ATS} & No attack &  0.84 & 70.3\% & -\\
                                      & Random Patch &  0.83 & 69.8\% & -2\%\\
                                      & TAP &  0.76 & 0.1\% & -17\%\\
                                      & NTAP &  0.61 & 0.1\% & -50\%\\
                                      \rowcolor{pink}
                                      & SlowFormer (ours) &  1.0 & 1.2\% & 35\%\\

        \midrule
                                &  \textbf{ViT-Small}            & \textbf{4.6}  & - & -\\
                                \cline{2-5}
        \multirow{5}{*}{ATS} & No attack &  3.1 & 79.2\% & - \\
                                      & Random Patch &  3.1 & 78.6\% & -1\%\\
                                      & TAP &  3.0 & 0.1\% & -7\%\\
                                      & NTAP &  2.4 & 0.1\% & -47\%\\
                                      \rowcolor{pink}
                                      & SlowFormer (ours) &  4.0 & 1.0\% & 60\%\\
        \midrule
                                &  \textbf{ViT-Base}            & \textbf{17.6}  & -  & -\\
                                \cline{2-5}
        \multirow{5}{*}{ATS} & No attack &  12.6 & 81.3\% & -\\
                                      & Random Patch &  12.5 & 81.2\% & -2\%\\
                                      & TAP &  12.0 & 0.1\% & -12\%\\
                                      & NTAP &  11.0 & 0.1\% & -32\%\\
                                      \rowcolor{pink}
                                      & SlowFormer (ours) &  15.4 & 0.2\% & 52\%\\
       \midrule
                                &  \textbf{ViT-Small}            & \textbf{4.6}  & -  & -\\
                                \cline{2-5}
        \multirow{5}{*}{AdaViT} & No attack    & 2.25 & 77.3\%  & - \\
                                & Random Patch  &     2.20     & 76.9\%  & -2\%\\
                                & TAP           &    2.28      &     0.1\%  & 1\%\\
                                & NTAP          &     2.15     & 0.1\% & -4\%\\
                                \rowcolor{pink}
                                & \ours         &      3.2     & 0.4\% & 40\%\\
       
        \bottomrule
        
        \end{tabular}
        }
    \end{minipage}
    \quad
    \begin{minipage}[t]{0.45\linewidth}
        \captionof{figure}{\textbf{Visualization of our Energy Attack on Vision Transformers:} We visualize the A-ViT-Small with and without our attack. We use patch size of $32$ for the attack (on the top-left corner). We show pruned tokens at layer $8$ of A-ViT-Small. Our attack can recover most of the pruned tokens, resulting in increased computation and power consumption. Note that although the patch is reasonably small and is in the corner of the view, it can affect the whole computational flow of the network. This is probably due to the global attention mechanism in transformers.}
        \label{fig:vis1}
        \centering
            \includegraphics[width=1.00\linewidth]{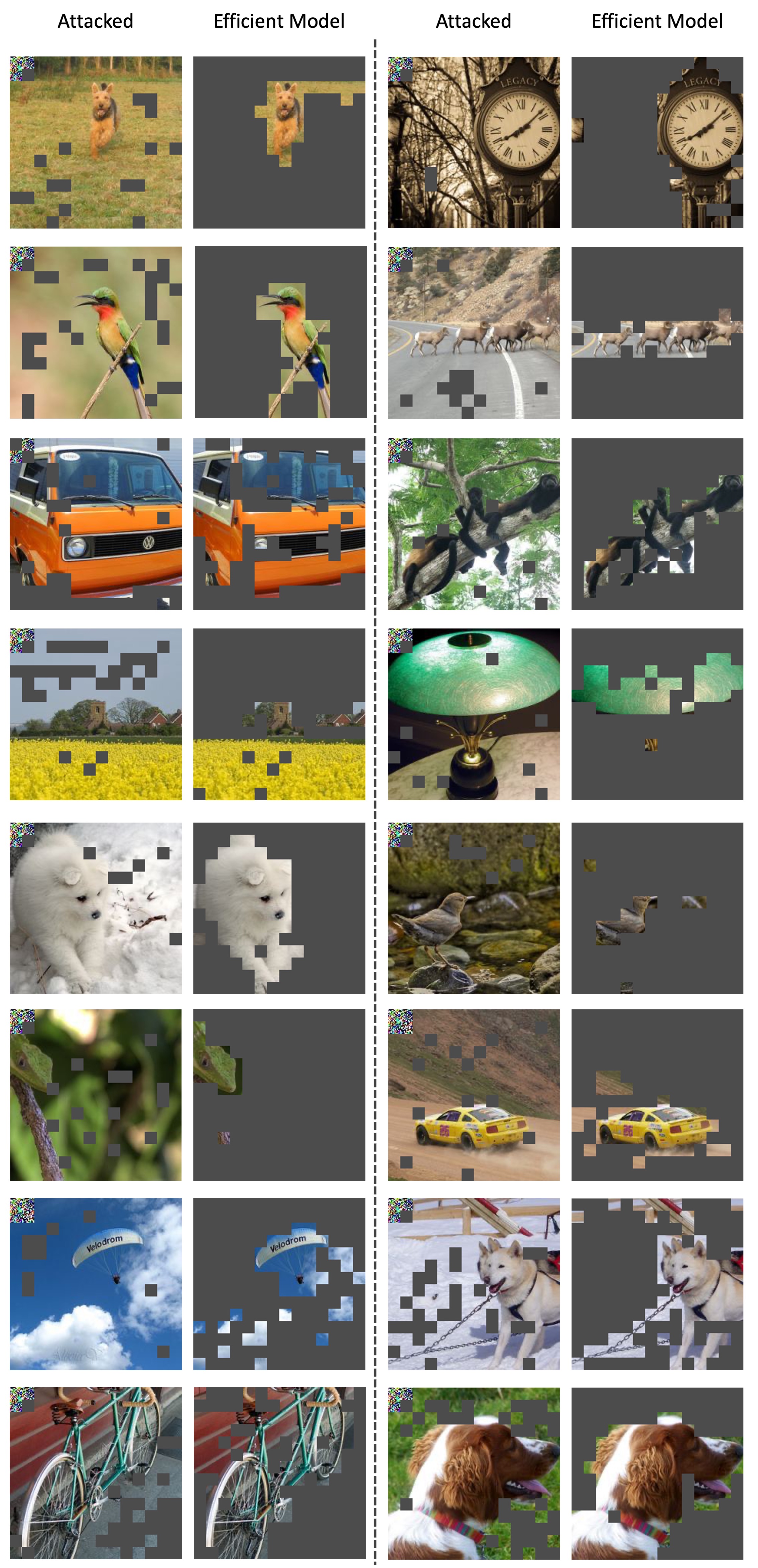}
        
    \end{minipage}
    
\end{table}

\noindent \textbf{Baselines:} We propose three alternative approaches to \sfo (ours) to generate the patch.

\noindent\textbf{Random Patch:} A simple baseline is to generate a randomly initialized patch. We sample IID pixel values from a uniform distribution between $0$ and $255$ to create the patch.  

\noindent\textbf{NTAP:} We consider a standard adversarial patch that is trained to attack the model task performance instead of compute. We use a non-targeted universal adversarial patch (NTAP) to attack the model. We train the patch to fool the model by misclassifying the image it is pasted on. We use the negative of the cross-entropy loss with the predicted and ground-truth labels as the loss to optimize the patch.

\noindent\textbf{TAP:} For this baseline, we train a universal targeted adversarial patch (TAP). 
The patch is optimized to classify all images in the train set to a single fixed category. Similar to NTAP, the adversarial attack here is on task performance and not computation. 
We experiment with ten randomly generated target category labels and report the averaged metrics.\\

\noindent \textbf{Implementation Details:}
We use PyTorch \citep{paszke2019pytorch} for all experiments. Unless specified, we use a patch of size $64 \times 64$, train and test on $224 \times 224$ images, and we paste the patch on the top-left corner. Note that our patch occupies just $8\%$ of the total area of an input image. We use AdamW \citep{loshchilov2017decoupled} optimizer to optimize the patches and use 4 NVIDIA RTX $3090$ GPUs for each experiment. We use varying batch sizes and learning rates for each of the computation-efficient methods. 

\noindent \textbf{ATS Details:} For our experiments on ATS, we use the weights of the DeiT model and replace regular attention blocks with the ATS block without training. As in ATS~\citep{fayyaz2022adaptive}, we replace layers 3 through 9 with the ATS block and set the maximum limit for the number of tokens sampled to $197$ for each layer. We train the patch for 2 epochs with a learning rate of $0.4$ for ViT-Tiny and $lr=0.2$ for ViT-Base and ViT-Small. 

\noindent \textbf{A-ViT Details:} We show results on ViT-Tiny and ViT-Small architectures using pretrained models provided by the authors of \avit. The patches are optimized for one epoch with a learning rate of $0.2$ and batch size of $128$. For the training of adversarial defense, we generate 5 patches per epoch of adversarial training and limit the number of iterations for patch generation to $500$. 

\noindent \textbf{AdaViT Details:} The authors of AdaViT provide a DeiT-S model pre-trained with their method. For this architecture, we freeze the weights and optimize for our adversarial patch. We use a $lr=0.2$ and a batch size of $128$. Additional details on all three efficient approaches are provided in the supplementary material.\\

\noindent \textbf{Results.} 
The results of our attack, \sfo, on various methods on ImageNet dataset are shown in table~\ref{tab:imagenet_attack}.
In A-ViT, we successfully recover $100\%$ of the computation reduced by \avit. Our attack has an Attack Success of $60\%$ on ATS and $40\%$ on AdaViT with ViT-Small. A random patch attack has little effect on both the accuracy and computation of the method. Both standard adversarial attack baselines, TAP and NTAP, reduce the accuracy to nearly $0\%$. Interestingly, these patches further decrease the computation of the efficient model being attacked. 
This might be because of the increased importance of adversarial patch tokens to the task and thus reduced importance of other tokens. Targeted patch (TAP) has a significant reduction in FLOPs on the ATS method. Since the token dropping in ATS relies on the distribution of attention values of classification tokens, a sharper distribution due to the increased importance of a token can result in a reduction in computation.

We report the results on CIFAR-10 dataset in Table~\ref{tab:cifar}. The efficient model (\avit) drastically reduces the computation from $1.26$ GFLOPs to $0.11$ GFLOPs. Most of the tokens are dropped as early as layer two in the efficient model. \sfo is able to effectively attack even in such extreme scenarios, achieving an Attack Success of $40\%$ and increasing the mean depth of tokens from nearly one to five. 

We additionally visualize the effectiveness of our attack in Figure~\ref{fig:vis1}. The un-attacked efficient method retains only
highly relevant tokens at the latter layers of the network. However, our attack results in nearly the entire image being passed
through all layers of the model for all inputs. In Fig.~\ref{fig:vis1}, we visualize the optimized patches for each of the three efficient
methods. 

\begin{table}[!tb]
    \begin{minipage}[t]{.45\linewidth}
        \caption{\textbf{Results on CIFAR10 dataset.} We report results on CIFAR10 dataset to show that our attack is not specific to ImageNet alone. CIFAR-10 is a small dataset compared to ImageNet and thus results in an extremely efficient \avit model. Our attack increases the FLOPs from 0.11 to 0.58 which restores nearly 41\% of the original reduction in the FLOPs.}
        \label{tab:cifar}
      \centering
        \scalebox{0.85}{
        \begin{tabular}{lccccc}
            \toprule
            Method & Model & Top-$1$ & Attack \\
            & FLOPs & Acc & Success \\
            \midrule
            ViT-Tiny   & 1.26  & - & -\\
            A-ViT-Tiny &  0.11 & 95.8\% & - \\
            \ours       & 0.58  & 60.2\% & 41\%\\
            \bottomrule
        \end{tabular}
        }
    \end{minipage} \quad
    \begin{minipage}[t]{.55\linewidth}
        \caption{\textbf{Accuracy controlled compute adversarial attack:} We attack the 
        the efficiency of A-ViT while either maintaining or destroying its classification performance. We observe that our attack can achieve a huge variation in task performance without affecting the Attack Success. The ability to attack the computation without affecting the task performance might be crucial in some applications. 
        }
        \label{tab:attack_acc}
        \centering
        \scalebox{0.85}{
        \begin{tabular}{lccccc} %
        \toprule
            Attack &  Model  & Attack & Top-$1$\\
                   &  GFLOPs & Success & Acc\\
            \midrule
            ViT-Tiny       & 1.26 & - & -\\
            \midrule
            No attack     & 0.87 &  -    & 71.4\%\\
            Acc agnostic   & 1.26 & 100\% & 4.7\%\\
            Preserve acc   & 1.23 & 92\%  & 68.5\% \\
            Destroy acc    & 1.26 & 100\% & 0.1\%  \\
        \bottomrule
        \end{tabular}
        }
    \end{minipage}%
\end{table}

\vspace{-.05in}
\subsection{Ablations:}
We perform all ablations on the \avit approach using their pretrained ViT-Tiny architecture model.

\noindent \textbf{Accuracy controlled compute adversarial attack:} As we show in Table~\ref{tab:imagenet_attack}, our attack can not only increase the computation, but also reduce the model accuracy. This can be desirable or hurtful based on the attacker's goals. A low-accuracy model might be an added benefit, similar to regular adversaries, but might also lead to the victim detecting the attack. Here, we show that it is possible to attack the computation of the model while either preserving or destroying the task performance by additionally employing a task loss in the patch optimization. As seen in Table~\ref{tab:attack_acc}, the accuracy can be significantly modified while maintaining a high Attack Success.

\noindent \textbf{Effect of patch size:} We vary the patch size from $64\times64$ to $16\times16$ (just a single token) and report the results in Table~\ref{tab:patch_size_ablation}. Interestingly, our attack with ViT-Small has a $73\%$ Attack Success with a $32\times32$ patch size, which occupies only $2\%$ of the input image area. 

\begin{table}[]
    \begin{minipage}[t]{0.5\linewidth}
        \caption{\textbf{Effect of patch size:} Analysis of the effect of adversarial patch
        size on the attack success rate on A-ViT. Our attach is reasonably successful even using a small patch size ($32\times32$), which is only 2\% of the image area. Interestingly, a small patch on the corner of the view affects the computational flow of the entire transformer model. This might be due to the global attention mechanism in transformers.
        }
        \label{tab:patch_size_ablation}
        
        \begin{tabular}{lcccccc} %
        \toprule
            Patch Size  &  Model & Top-1 & Attack \\
            (Area) & GFLOPs  & Accuracy & Success\\
            \midrule
            ViT-Tiny   & 1.26 &  - & -\\
            A-ViT-Tiny & 0.87 & 71.4\% & -\\
            \midrule
            64 (8\%)   & 1.26  & 4.7\%  & 100\% \\
            48 (5\%)   & 1.26  & 1.8\%  & 99\%  \\
            32 (2\%)   & 1.22  & 17.4\% & 90\%  \\
            16 (0.5\%) & 0.98   & 63.3\% & 27\%\\
            \midrule
            ViT-Small & 4.6  & - & -\\
            A-ViT-Small & 3.7  & 78.8\% & -\\
            \midrule
            64 (8\%)   & 4.6   & 2.3\%  & 99\% \\
            48 (5\%)   & 4.6   & 5.1\%  & 98\% \\
            32 (2\%)   & 4.4   & 39.5\% & 78\% \\
            16 (0.5\%) & 3.8   & 78.2\% & 16\%\\
        \bottomrule
        \end{tabular}
    \end{minipage}
    \quad
    \vspace{-.1in}
    \begin{minipage}[t]{0.45\linewidth}
    \centering
    \caption{\textbf{Defense using adversarial training:} We propose and show the impact of our defense for our adversarial attack on \avit. Our defense is simply maintaining a set of universal patches and training the model to be robust to a random sample of those at each iteration. The defense reduces the computation to some extent (1.26 to 1.01), but is still far from the original unattacked model (0.87).}
    \label{tab:defense}
    \scalebox{0.90}
    {
    \begin{tabular}{lccc}
        \toprule
        Method & GFLOPs & Top-1 & Attack \\
        & & Acc. & Success \\
        \midrule
        No attack & 0.87 & 71.4 & -\\
        \sfo   & 1.26 & 4.7\%  & 100\%\\
        \multirow{2}{*}{\parbox{2.0cm}{Adv Defense + \sfo}}   & \multirow{2}{*}{1.01} & \multirow{2}{*}{65.8\%} & 
        \multirow{2}{*}{34\%}\\
        \\
        \bottomrule
    \end{tabular}
    }
\includegraphics[width=1.0\textwidth]{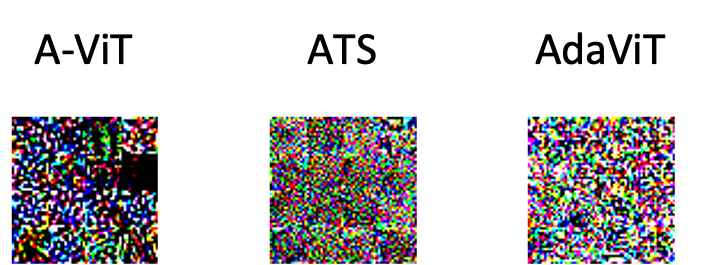}
\captionof{figure}{\textbf{Visualization of optimized patch:} We show the learned universal patches for
    each of the three efficient methods.} 
\label{fig:vis1}
    \end{minipage}
    
\end{table}

\noindent \textbf{Effect of patch location:} We vary the location of the patch to study the effect of location on the Attack Success. 
We randomly sample a location in the image to paste the patch on. We perform five such experiments and observe an Attack Success of $100\%$ for all patch locations. 

\vspace{-.05in}
\subsection{Adversarial training based defense}
Our simple defense that is adopted from standard adversarial training is explained in Section \ref{sec:defense}. The results for defending against attacking \avit are shown in Table \ref{tab:defense}. The original \avit reduces the GFLOPs from 1.26 to 0.87, our attack increases it back to 1.26 with 100\% attack success. The proposed defense reduces the GFLOPs to 1.01 which is still higher than the original 0.87. We hope our paper encourages the community to develop better defense methods to reduce the vulnerability of efficient vision transformers.

    
    

\vspace{-.1in}
\section{Conclusion}
Recently, we have seen efficient transfer models in which the computation is adaptively modified based on the input. We argue that this is an important research direction and that there will be more progress in this direction in the future. However, we show that the current methods are vulnerable to a universal adversarial patch that increases the computation and thus power consumption at inference time. Our experiments show promising results for three SOTA efficient transformer models, where a small patch that is optimized on the training data can increase the computation to the maximum possible level in the testing data in some settings. We also propose a defense that reduces the effectiveness of our attack. We hope our paper will encourage the community to study such attacks and develop better defense methods.

{\bf Acknowledgment:}
This work was partially supported by the Defense Advanced
Research Projects Agency (DARPA) under Contract No. HR00112190135 and funding from
NSF grant 1845216. Any opinions, findings, conclusions, or recommendations expressed in this paper are those of the authors and do not necessarily reflect the views of the funding agencies.

\FloatBarrier


\bibliography{egbib}

\begin{thebibliography}{91}
\providecommand{\natexlab}[1]{#1}
\providecommand{\url}[1]{\texttt{#1}}
\expandafter\ifx\csname urlstyle\endcsname\relax
  \providecommand{\doi}[1]{doi: #1}\else
  \providecommand{\doi}{doi: \begingroup \urlstyle{rm}\Url}\fi

\bibitem[sta()]{starship}
Starship robot.
\newblock
  \url{https://www.wevolver.com/specs/starship-technologies-starship-robot}.

\bibitem[Ali et~al.(2021)Ali, Touvron, Caron, Bojanowski, Douze, Joulin,
  Laptev, Neverova, Synnaeve, Verbeek, et~al.]{ali2021xcit}
Alaaeldin Ali, Hugo Touvron, Mathilde Caron, Piotr Bojanowski, Matthijs Douze,
  Armand Joulin, Ivan Laptev, Natalia Neverova, Gabriel Synnaeve, Jakob
  Verbeek, et~al.
\newblock Xcit: Cross-covariance image transformers.
\newblock \emph{Advances in neural information processing systems}, 34, 2021.

\bibitem[Bejnordi et~al.(2019)Bejnordi, Blankevoort, and
  Welling]{bejnordi2019batch}
Babak~Ehteshami Bejnordi, Tijmen Blankevoort, and Max Welling.
\newblock Batch-shaping for learning conditional channel gated networks.
\newblock \emph{arXiv preprint arXiv:1907.06627}, 2019.

\bibitem[Bolukbasi et~al.(2017)Bolukbasi, Wang, Dekel, and
  Saligrama]{bolukbasi2017adaptive}
Tolga Bolukbasi, Joseph Wang, Ofer Dekel, and Venkatesh Saligrama.
\newblock Adaptive neural networks for efficient inference.
\newblock In \emph{International Conference on Machine Learning}, pp.\
  527--536. PMLR, 2017.

\bibitem[Bolya et~al.(2022)Bolya, Fu, Dai, Zhang, Feichtenhofer, and
  Hoffman]{bolya2022token}
Daniel Bolya, Cheng-Yang Fu, Xiaoliang Dai, Peizhao Zhang, Christoph
  Feichtenhofer, and Judy Hoffman.
\newblock Token merging: Your vit but faster.
\newblock \emph{arXiv preprint arXiv:2210.09461}, 2022.

\bibitem[Brown et~al.(2022)Brown, Zhao, Shumailov, and
  Mullins]{brown2022dartformer}
Jason~Ross Brown, Yiren Zhao, Ilia Shumailov, and Robert~D Mullins.
\newblock Dartformer: Finding the best type of attention.
\newblock \emph{arXiv preprint arXiv:2210.00641}, 2022.

\bibitem[Brown et~al.(2017)Brown, Man{\'e}, Roy, Abadi, and
  Gilmer]{brown2017adversarial}
Tom~B Brown, Dandelion Man{\'e}, Aurko Roy, Mart{\'\i}n Abadi, and Justin
  Gilmer.
\newblock Adversarial patch.
\newblock \emph{arXiv preprint arXiv:1712.09665}, 2017.

\bibitem[Cai et~al.(2021)Cai, Shu, and Wang]{cai2021dynamic}
Shaofeng Cai, Yao Shu, and Wei Wang.
\newblock Dynamic routing networks.
\newblock In \emph{Proceedings of the IEEE/CVF Winter Conference on
  Applications of Computer Vision}, pp.\  3588--3597, 2021.

\bibitem[Cao et~al.(2019)Cao, Ma, Xiao, Zhang, Liu, Zhang, Nie, and
  Yang]{cao2019seernet}
Shijie Cao, Lingxiao Ma, Wencong Xiao, Chen Zhang, Yunxin Liu, Lintao Zhang,
  Lanshun Nie, and Zhi Yang.
\newblock Seernet: Predicting convolutional neural network feature-map sparsity
  through low-bit quantization.
\newblock In \emph{Proceedings of the IEEE/CVF Conference on Computer Vision
  and Pattern Recognition}, pp.\  11216--11225, 2019.

\bibitem[Carion et~al.(2020)Carion, Massa, Synnaeve, Usunier, Kirillov, and
  Zagoruyko]{detr}
Nicolas Carion, Francisco Massa, Gabriel Synnaeve, Nicolas Usunier, Alexander
  Kirillov, and Sergey Zagoruyko.
\newblock End-to-end object detection with transformers.
\newblock In \emph{European Conference on Computer Vision (ECCV)}, 2020.

\bibitem[Caron et~al.(2021)Caron, Touvron, Misra, J{\'e}gou, Mairal,
  Bojanowski, and Joulin]{caron2021emerging}
Mathilde Caron, Hugo Touvron, Ishan Misra, Herv{\'e} J{\'e}gou, Julien Mairal,
  Piotr Bojanowski, and Armand Joulin.
\newblock Emerging properties in self-supervised vision transformers.
\newblock In \emph{Proceedings of the IEEE/CVF International Conference on
  Computer Vision}, pp.\  9650--9660, 2021.

\bibitem[Chen et~al.(2019)Chen, Lin, Ren, Lu, and Zhou]{chen2019self}
Guangyi Chen, Chunze Lin, Liangliang Ren, Jiwen Lu, and Jie Zhou.
\newblock Self-critical attention learning for person re-identification.
\newblock In \emph{ICCV}, pp.\  9637--9646, 2019.

\bibitem[Chen et~al.(2021{\natexlab{a}})Chen, Wang, Guo, Zhang, and
  Sun]{chen2021dynamic}
Jin Chen, Xijun Wang, Zichao Guo, Xiangyu Zhang, and Jian Sun.
\newblock Dynamic region-aware convolution.
\newblock In \emph{Proceedings of the IEEE/CVF Conference on Computer Vision
  and Pattern Recognition}, pp.\  8064--8073, 2021{\natexlab{a}}.

\bibitem[Chen et~al.()Chen, Chen, Haque, Liu, and Yang]{chenslothbomb}
Simin Chen, Hanlin Chen, Mirazul Haque, Cong Liu, and Wei Yang.
\newblock Slothbomb: Efficiency poisoning attack against dynamic neural
  networks.

\bibitem[Chen et~al.(2021{\natexlab{b}})Chen, Haque, Song, Liu, and
  Yang]{chen2021transslowdown}
Simin Chen, Mirazul Haque, Zihe Song, Cong Liu, and Wei Yang.
\newblock Transslowdown: Efficiency attacks on neural machine translation
  systems.
\newblock 2021{\natexlab{b}}.

\bibitem[Chen et~al.(2022)Chen, Song, Haque, Liu, and
  Yang]{chen2022nicgslowdown}
Simin Chen, Zihe Song, Mirazul Haque, Cong Liu, and Wei Yang.
\newblock Nicgslowdown: Evaluating the efficiency robustness of neural image
  caption generation models.
\newblock In \emph{Proceedings of the IEEE/CVF Conference on Computer Vision
  and Pattern Recognition}, pp.\  15365--15374, 2022.

\bibitem[Cheng et~al.(2021)Cheng, Schwing, and Kirillov]{maskformer}
Bowen Cheng, Alexander~G. Schwing, and Alexander Kirillov.
\newblock Per-pixel classification is not all you need for semantic
  segmentation.
\newblock In \emph{Advances in Neural Information Processing Systems
  (NeurIPS)}, 2021.

\bibitem[Deng et~al.(2009)Deng, Dong, Socher, Li, Li, and
  Fei-Fei]{deng2009imagenet}
Jia Deng, Wei Dong, Richard Socher, Li-Jia Li, Kai Li, and Li~Fei-Fei.
\newblock Imagenet: A large-scale hierarchical image database.
\newblock In \emph{2009 IEEE conference on computer vision and pattern
  recognition}, pp.\  248--255. Ieee, 2009.

\bibitem[Dosovitskiy et~al.(2020)Dosovitskiy, Beyer, Kolesnikov, Weissenborn,
  Zhai, Unterthiner, Dehghani, Minderer, Heigold, Gelly,
  et~al.]{dosovitskiy2020image}
Alexey Dosovitskiy, Lucas Beyer, Alexander Kolesnikov, Dirk Weissenborn,
  Xiaohua Zhai, Thomas Unterthiner, Mostafa Dehghani, Matthias Minderer, Georg
  Heigold, Sylvain Gelly, et~al.
\newblock An image is worth 16x16 words: Transformers for image recognition at
  scale.
\newblock \emph{arXiv preprint arXiv:2010.11929}, 2020.

\bibitem[Dosovitskiy et~al.(2021)Dosovitskiy, Beyer, Kolesnikov, Weissenborn,
  Zhai, Unterthiner, Dehghani, Minderer, Heigold, Gelly, Uszkoreit, and
  Houlsby]{vit}
Alexey Dosovitskiy, Lucas Beyer, Alexander Kolesnikov, Dirk Weissenborn,
  Xiaohua Zhai, Thomas Unterthiner, Mostafa Dehghani, Matthias Minderer, Georg
  Heigold, Sylvain Gelly, Jakob Uszkoreit, and Neil Houlsby.
\newblock An image is worth 16x16 words: Transformers for image recognition at
  scale.
\newblock In \emph{International Conference on Learning Representations
  (ICLR)}, 2021.

\bibitem[Elbayad et~al.(2019)Elbayad, Gu, Grave, and Auli]{elbayad2019depth}
Maha Elbayad, Jiatao Gu, Edouard Grave, and Michael Auli.
\newblock Depth-adaptive transformer.
\newblock \emph{arXiv preprint arXiv:1910.10073}, 2019.

\bibitem[Fan et~al.(2019)Fan, Chen, Kuehne, Pistoia, and Cox]{fan2019blvnet}
Quanfu Fan, Chun-Fu~(Ricarhd) Chen, Hilde Kuehne, Marco Pistoia, and David Cox.
\newblock {More Is Less: Learning Efficient Video Representations by Temporal
  Aggregation Modules}.
\newblock In \emph{Advances in Neural Information Processing Systems
  (NeurIPS)}, 2019.

\bibitem[Fayyaz et~al.(2021)Fayyaz, Koohpayegani, Jafari, Sengupta, Joze,
  Sommerlade, Pirsiavash, and Gall]{ats}
Mohsen Fayyaz, Soroush~Abbasi Koohpayegani, Farnoush~Rezaei Jafari, Sunando
  Sengupta, Hamid Reza~Vaezi Joze, Eric Sommerlade, Hamed Pirsiavash, and
  Juergen Gall.
\newblock Adaptive token sampling for efficient vision transformers, 2021.
\newblock URL \url{https://arxiv.org/abs/2111.15667}.

\bibitem[Fayyaz et~al.(2022)Fayyaz, Koohpayegani, Jafari, Sengupta, Joze,
  Sommerlade, Pirsiavash, and Gall]{fayyaz2022adaptive}
Mohsen Fayyaz, Soroush~Abbasi Koohpayegani, Farnoush~Rezaei Jafari, Sunando
  Sengupta, Hamid Reza~Vaezi Joze, Eric Sommerlade, Hamed Pirsiavash, and
  J{\"u}rgen Gall.
\newblock Adaptive token sampling for efficient vision transformers.
\newblock In \emph{Computer Vision--ECCV 2022: 17th European Conference, Tel
  Aviv, Israel, October 23--27, 2022, Proceedings, Part XI}, pp.\  396--414.
  Springer, 2022.

\bibitem[Fedus et~al.(2021)Fedus, Zoph, and Shazeer]{fedus2021switch}
William Fedus, Barret Zoph, and Noam Shazeer.
\newblock Switch transformers: Scaling to trillion parameter models with simple
  and efficient sparsity.
\newblock \emph{J. Mach. Learn. Res}, 23:\penalty0 1--40, 2021.

\bibitem[Feinman et~al.(2017)Feinman, Curtin, Shintre, and
  Gardner]{feinman2017detecting}
Reuben Feinman, Ryan~R Curtin, Saurabh Shintre, and Andrew~B Gardner.
\newblock Detecting adversarial samples from artifacts.
\newblock \emph{arXiv preprint arXiv:1703.00410}, 2017.

\bibitem[Figurnov et~al.(2017)Figurnov, Collins, Zhu, Zhang, Huang, Vetrov, and
  Salakhutdinov]{figurnov2017spatially}
Michael Figurnov, Maxwell~D Collins, Yukun Zhu, Li~Zhang, Jonathan Huang,
  Dmitry Vetrov, and Ruslan Salakhutdinov.
\newblock Spatially adaptive computation time for residual networks.
\newblock In \emph{Proceedings of the IEEE conference on computer vision and
  pattern recognition}, pp.\  1039--1048, 2017.

\bibitem[Fu et~al.(2017)Fu, Zheng, and Mei]{fu2017look}
Jianlong Fu, Heliang Zheng, and Tao Mei.
\newblock Look closer to see better: Recurrent attention convolutional neural
  network for fine-grained image recognition.
\newblock In \emph{IEEE Conference on Computer Vision and Pattern Recognition
  (CVPR)}, 2017.

\bibitem[Fu et~al.(2022)Fu, Zhang, Wu, Wan, and Lin]{fu2022patch}
Yonggan Fu, Shunyao Zhang, Shang Wu, Cheng Wan, and Yingyan Lin.
\newblock Patch-fool: Are vision transformers always robust against adversarial
  perturbations?
\newblock \emph{arXiv preprint arXiv:2203.08392}, 2022.

\bibitem[Gao et~al.(2018)Gao, Zhao, Dudziak, Mullins, and Xu]{gao2018dynamic}
Xitong Gao, Yiren Zhao, {\L}ukasz Dudziak, Robert Mullins, and Cheng-zhong Xu.
\newblock Dynamic channel pruning: Feature boosting and suppression.
\newblock \emph{arXiv preprint arXiv:1810.05331}, 2018.

\bibitem[Goodfellow et~al.(2014)Goodfellow, Shlens, and
  Szegedy]{goodfellow2014explaining}
Ian~J Goodfellow, Jonathon Shlens, and Christian Szegedy.
\newblock Explaining and harnessing adversarial examples.
\newblock In \emph{arXiv preprint arXiv:1412.6572}, 2014.

\bibitem[Graves(2016)]{graves2016adaptive}
Alex Graves.
\newblock Adaptive computation time for recurrent neural networks.
\newblock \emph{arXiv preprint arXiv:1603.08983}, 2016.

\bibitem[Guan et~al.(2017)Guan, Liu, Liu, and Peng]{guan2017energy}
Jiaqi Guan, Yang Liu, Qiang Liu, and Jian Peng.
\newblock Energy-efficient amortized inference with cascaded deep classifiers.
\newblock \emph{arXiv preprint arXiv:1710.03368}, 2017.

\bibitem[Haque et~al.(2020)Haque, Chauhan, Liu, and Yang]{haque2020ilfo}
Mirazul Haque, Anki Chauhan, Cong Liu, and Wei Yang.
\newblock Ilfo: Adversarial attack on adaptive neural networks.
\newblock In \emph{Proceedings of the IEEE/CVF Conference on Computer Vision
  and Pattern Recognition}, pp.\  14264--14273, 2020.

\bibitem[Haque et~al.(2021)Haque, Chen, Haque, Liu, and
  Yang]{haque2021nodeattack}
Mirazul Haque, Simin Chen, Wasif~Arman Haque, Cong Liu, and Wei Yang.
\newblock Nodeattack: Adversarial attack on the energy consumption of neural
  odes.
\newblock 2021.

\bibitem[He et~al.(2021)He, Chen, Xie, Li, Doll{\'a}r, and
  Girshick]{he2021masked}
Kaiming He, Xinlei Chen, Saining Xie, Yanghao Li, Piotr Doll{\'a}r, and Ross
  Girshick.
\newblock Masked autoencoders are scalable vision learners.
\newblock \emph{arXiv preprint arXiv:2111.06377}, 2021.

\bibitem[Herrmann et~al.(2020)Herrmann, Bowen, and Zabih]{herrmann2020channel}
Charles Herrmann, Richard~Strong Bowen, and Ramin Zabih.
\newblock Channel selection using gumbel softmax.
\newblock In \emph{Computer Vision--ECCV 2020: 16th European Conference,
  Glasgow, UK, August 23--28, 2020, Proceedings, Part XXVII}, pp.\  241--257.
  Springer, 2020.

\bibitem[Hinton et~al.(2015)Hinton, Vinyals, and Dean]{hinton2015distilling}
Geoffrey Hinton, Oriol Vinyals, and Jeff Dean.
\newblock Distilling the knowledge in a neural network.
\newblock \emph{arXiv preprint arXiv:1503.02531}, 2015.

\bibitem[Ho et~al.(2019)Ho, Kalchbrenner, Weissenborn, and
  Salimans]{ho2019axial}
Jonathan Ho, Nal Kalchbrenner, Dirk Weissenborn, and Tim Salimans.
\newblock Axial attention in multidimensional transformers.
\newblock \emph{arXiv preprint arXiv:1912.12180}, 2019.

\bibitem[Hong et~al.(2020)Hong, Kaya, Modoranu, and
  Dumitra{\c{s}}]{hong2020panda}
Sanghyun Hong, Yi{\u{g}}itcan Kaya, Ionu{\c{t}}-Vlad Modoranu, and Tudor
  Dumitra{\c{s}}.
\newblock A panda? no, it's a sloth: Slowdown attacks on adaptive multi-exit
  neural network inference.
\newblock \emph{arXiv preprint arXiv:2010.02432}, 2020.

\bibitem[Howard et~al.(2017)Howard, Zhu, Chen, Kalenichenko, Wang, Weyand,
  Andreetto, and Adam]{howard2017mobilenets}
Andrew~G Howard, Menglong Zhu, Bo~Chen, Dmitry Kalenichenko, Weijun Wang,
  Tobias Weyand, Marco Andreetto, and Hartwig Adam.
\newblock Mobilenets: Efficient convolutional neural networks for mobile vision
  applications.
\newblock \emph{arXiv preprint arXiv:1704.04861}, 2017.

\bibitem[Hua et~al.(2019)Hua, Zhou, De~Sa, Zhang, and Suh]{hua2019channel}
Weizhe Hua, Yuan Zhou, Christopher~M De~Sa, Zhiru Zhang, and G~Edward Suh.
\newblock Channel gating neural networks.
\newblock \emph{Advances in Neural Information Processing Systems}, 32, 2019.

\bibitem[Huang et~al.(2017)Huang, Chen, Li, Wu, Van Der~Maaten, and
  Weinberger]{huang2017multi}
Gao Huang, Danlu Chen, Tianhong Li, Felix Wu, Laurens Van Der~Maaten, and
  Kilian~Q Weinberger.
\newblock Multi-scale dense networks for resource efficient image
  classification.
\newblock \emph{arXiv preprint arXiv:1703.09844}, 2017.

\bibitem[Katharopoulos et~al.(2020)Katharopoulos, Vyas, Pappas, and
  Fleuret]{katharopoulos2020transformers}
Angelos Katharopoulos, Apoorv Vyas, Nikolaos Pappas, and Fran{\c{c}}ois
  Fleuret.
\newblock Transformers are rnns: Fast autoregressive transformers with linear
  attention.
\newblock In \emph{International Conference on Machine Learning}, pp.\
  5156--5165. PMLR, 2020.

\bibitem[Keles et~al.(2022)Keles, Wijewardena, and
  Hegde]{keles2022computational}
Feyza~Duman Keles, Pruthuvi~Mahesakya Wijewardena, and Chinmay Hegde.
\newblock On the computational complexity of self-attention.
\newblock \emph{arXiv preprint arXiv:2209.04881}, 2022.

\bibitem[Kong \& Fowlkes(2019)Kong and Fowlkes]{kong2019pixel}
Shu Kong and Charless Fowlkes.
\newblock Pixel-wise attentional gating for scene parsing.
\newblock In \emph{2019 IEEE Winter Conference on Applications of Computer
  Vision (WACV)}, pp.\  1024--1033. IEEE, 2019.

\bibitem[Koohpayegani \& Pirsiavash(2022)Koohpayegani and
  Pirsiavash]{koohpayegani2022sima}
Soroush~Abbasi Koohpayegani and Hamed Pirsiavash.
\newblock Sima: Simple softmax-free attention for vision transformers.
\newblock \emph{arXiv preprint arXiv:2206.08898}, 2022.

\bibitem[Krizhevsky(2009)]{cifar}
Alex Krizhevsky.
\newblock Learning multiple layers of features from tiny images.
\newblock Technical report, University of Toronto, 2009.

\bibitem[Kurakin et~al.(2018)Kurakin, Goodfellow, and
  Bengio]{kurakin2018adversarial}
Alexey Kurakin, Ian~J Goodfellow, and Samy Bengio.
\newblock Adversarial examples in the physical world.
\newblock In \emph{Artificial intelligence safety and security}, pp.\  99--112.
  Chapman and Hall/CRC, 2018.

\bibitem[Li et~al.(2016)Li, Kadav, Durdanovic, Samet, and Graf]{li2016pruning}
Hao Li, Asim Kadav, Igor Durdanovic, Hanan Samet, and Hans~Peter Graf.
\newblock Pruning filters for efficient convnets.
\newblock \emph{arXiv preprint arXiv:1608.08710}, 2016.

\bibitem[Li \& Li(2017)Li and Li]{li2017adversarial}
Xin Li and Fuxin Li.
\newblock Adversarial examples detection in deep networks with convolutional
  filter statistics.
\newblock In \emph{Proceedings of the IEEE international conference on computer
  vision}, pp.\  5764--5772, 2017.

\bibitem[Li et~al.(2017)Li, Yang, Liu, Zhou, Wen, and Xu]{li2017dynamic}
Zhichao Li, Yi~Yang, Xiao Liu, Feng Zhou, Shilei Wen, and Wei Xu.
\newblock Dynamic computational time for visual attention.
\newblock In \emph{Proceedings of the IEEE International Conference on Computer
  Vision Workshops}, pp.\  1199--1209, 2017.

\bibitem[Liu et~al.(2022)Liu, Pan, He, Cai, and Zhuang]{liu2022ecoformer}
Jing Liu, Zizheng Pan, Haoyu He, Jianfei Cai, and Bohan Zhuang.
\newblock Ecoformer: Energy-saving attention with linear complexity.
\newblock \emph{arXiv preprint arXiv:2209.09004}, 2022.

\bibitem[Liu et~al.(2021{\natexlab{a}})Liu, Lin, Cao, Hu, Wei, Zhang, Lin, and
  Guo]{liu2021swin}
Ze~Liu, Yutong Lin, Yue Cao, Han Hu, Yixuan Wei, Zheng Zhang, Stephen Lin, and
  Baining Guo.
\newblock Swin transformer: Hierarchical vision transformer using shifted
  windows.
\newblock In \emph{Proceedings of the IEEE/CVF International Conference on
  Computer Vision}, pp.\  10012--10022, 2021{\natexlab{a}}.

\bibitem[Liu et~al.(2021{\natexlab{b}})Liu, Lin, Cao, Hu, Wei, Zhang, Lin, and
  Guo]{swin}
Ze~Liu, Yutong Lin, Yue Cao, Han Hu, Yixuan Wei, Zheng Zhang, Stephen Lin, and
  Baining Guo.
\newblock Swin transformer: Hierarchical vision transformer using shifted
  windows.
\newblock In \emph{Proceedings of the IEEE/CVF International Conference on
  Computer Vision (ICCV)}, 2021{\natexlab{b}}.

\bibitem[Loshchilov \& Hutter(2019)Loshchilov and
  Hutter]{loshchilov2017decoupled}
Ilya Loshchilov and Frank Hutter.
\newblock Decoupled weight decay regularization.
\newblock 2019.

\bibitem[Lu et~al.(2021)Lu, Yao, Zhang, Zhu, Xu, Gao, Xu, Xiang, and
  Zhang]{lu2021soft}
Jiachen Lu, Jinghan Yao, Junge Zhang, Xiatian Zhu, Hang Xu, Weiguo Gao,
  Chunjing Xu, Tao Xiang, and Li~Zhang.
\newblock Soft: Softmax-free transformer with linear complexity.
\newblock \emph{Advances in Neural Information Processing Systems}, 34, 2021.

\bibitem[Lu et~al.(2020)Lu, Jiao, and Zhang]{lu2020twinbert}
Wenhao Lu, Jian Jiao, and Ruofei Zhang.
\newblock Twinbert: Distilling knowledge to twin-structured bert models for
  efficient retrieval.
\newblock \emph{arXiv preprint arXiv:2002.06275}, 2020.

\bibitem[Maddison et~al.(2016)Maddison, Mnih, and Teh]{maddison2016concrete}
Chris~J Maddison, Andriy Mnih, and Yee~Whye Teh.
\newblock The concrete distribution: A continuous relaxation of discrete random
  variables.
\newblock \emph{arXiv preprint arXiv:1611.00712}, 2016.

\bibitem[Marin et~al.(2021)Marin, Chang, Ranjan, Prabhu, Rastegari, and
  Tuzel]{marin2021token}
Dmitrii Marin, Jen-Hao~Rick Chang, Anurag Ranjan, Anish Prabhu, Mohammad
  Rastegari, and Oncel Tuzel.
\newblock Token pooling in vision transformers.
\newblock \emph{arXiv preprint arXiv:2110.03860}, 2021.

\bibitem[Meng et~al.(2022)Meng, Li, Chen, Lan, Wu, Jiang, and
  Lim]{meng2022adavit}
Lingchen Meng, Hengduo Li, Bor-Chun Chen, Shiyi Lan, Zuxuan Wu, Yu-Gang Jiang,
  and Ser-Nam Lim.
\newblock Adavit: Adaptive vision transformers for efficient image recognition.
\newblock In \emph{Proceedings of the IEEE/CVF Conference on Computer Vision
  and Pattern Recognition}, pp.\  12309--12318, 2022.

\bibitem[Pan et~al.(2022)Pan, Zheng, Fan, Rahmani, Ke, and
  Liu]{pan2022gradauto}
Jianhong Pan, Qichen Zheng, Zhipeng Fan, Hossein Rahmani, Qiuhong Ke, and Jun
  Liu.
\newblock Gradauto: Energy-oriented attack on dynamic neural networks.
\newblock In \emph{Computer Vision--ECCV 2022: 17th European Conference, Tel
  Aviv, Israel, October 23--27, 2022, Proceedings, Part IV}, pp.\  637--653.
  Springer, 2022.

\bibitem[Papernot et~al.(2016)Papernot, McDaniel, Wu, Jha, and
  Swami]{papernot2016distillation}
Nicolas Papernot, Patrick McDaniel, Xi~Wu, Somesh Jha, and Ananthram Swami.
\newblock Distillation as a defense to adversarial perturbations against deep
  neural networks.
\newblock In \emph{2016 IEEE symposium on security and privacy (SP)}, pp.\
  582--597. IEEE, 2016.

\bibitem[Paszke et~al.(2019)Paszke, Gross, Massa, Lerer, Bradbury, Chanan,
  Killeen, Lin, Gimelshein, Antiga, et~al.]{paszke2019pytorch}
Adam Paszke, Sam Gross, Francisco Massa, Adam Lerer, James Bradbury, Gregory
  Chanan, Trevor Killeen, Zeming Lin, Natalia Gimelshein, Luca Antiga, et~al.
\newblock Pytorch: An imperative style, high-performance deep learning library.
\newblock \emph{Advances in neural information processing systems}, 32, 2019.

\bibitem[Radford et~al.(2021)Radford, Kim, Hallacy, Ramesh, Goh, Agarwal,
  Sastry, Askell, Mishkin, Clark, Krueger, and Sutskever]{clip@clip}
Alec Radford, Jong~Wook Kim, Chris Hallacy, Aditya Ramesh, Gabriel Goh,
  Sandhini Agarwal, Girish Sastry, Amanda Askell, Pamela Mishkin, Jack Clark,
  Gretchen Krueger, and Ilya Sutskever.
\newblock Learning transferable visual models from natural language
  supervision, 2021.
\newblock URL \url{https://arxiv.org/abs/2103.00020}.

\bibitem[Rao et~al.(2021{\natexlab{a}})Rao, Zhao, Liu, Lu, Zhou, and
  Hsieh]{rao2021dynamicvit}
Yongming Rao, Wenliang Zhao, Benlin Liu, Jiwen Lu, Jie Zhou, and Cho-Jui Hsieh.
\newblock Dynamicvit: Efficient vision transformers with dynamic token
  sparsification.
\newblock \emph{Advances in neural information processing systems}, 34,
  2021{\natexlab{a}}.

\bibitem[Rao et~al.(2021{\natexlab{b}})Rao, Zhao, Zhu, Lu, and
  Zhou]{global_filter_net}
Yongming Rao, Wenliang Zhao, Zheng Zhu, Jiwen Lu, and Jie Zhou.
\newblock Global filter networks for image classification.
\newblock In \emph{Advances in Neural Information Processing Systems
  (NeurIPS)}, 2021{\natexlab{b}}.

\bibitem[Rastegari et~al.(2016)Rastegari, Ordonez, Redmon, and
  Farhadi]{rastegari2016xnornet}
Mohammad Rastegari, Vicente Ordonez, Joseph Redmon, and Ali Farhadi.
\newblock Xnor-net: Imagenet classification using binary convolutional neural
  networks, 2016.

\bibitem[Ren et~al.(2018)Ren, Pokrovsky, Yang, and Urtasun]{ren2018sbnet}
Mengye Ren, Andrei Pokrovsky, Bin Yang, and Raquel Urtasun.
\newblock Sbnet: Sparse blocks network for fast inference.
\newblock In \emph{Proceedings of the IEEE Conference on Computer Vision and
  Pattern Recognition}, pp.\  8711--8720, 2018.

\bibitem[Saha et~al.(2019)Saha, Subramanya, Patil, and
  Pirsiavash]{Saha2019AdversarialPE}
Aniruddha Saha, Akshayvarun Subramanya, Koninika~B. Patil, and Hamed
  Pirsiavash.
\newblock Adversarial patches exploiting contextual reasoning in object
  detection.
\newblock In \emph{ArXiv}, volume abs/1910.00068, 2019.

\bibitem[Shen et~al.(2021)Shen, Zhang, Zhao, Yi, and Li]{shen2021efficient}
Zhuoran Shen, Mingyuan Zhang, Haiyu Zhao, Shuai Yi, and Hongsheng Li.
\newblock Efficient attention: Attention with linear complexities.
\newblock In \emph{Proceedings of the IEEE/CVF Winter Conference on
  Applications of Computer Vision}, pp.\  3531--3539, 2021.

\bibitem[Szegedy et~al.(2013)Szegedy, Zaremba, Sutskever, Bruna, Erhan,
  Goodfellow, and Fergus]{szegedy2013intriguing}
Christian Szegedy, Wojciech Zaremba, Ilya Sutskever, Joan Bruna, Dumitru Erhan,
  Ian Goodfellow, and Rob Fergus.
\newblock Intriguing properties of neural networks.
\newblock \emph{arXiv preprint arXiv:1312.6199}, 2013.

\bibitem[Teerapittayanon et~al.(2016)Teerapittayanon, McDanel, and
  Kung]{teerapittayanon2016branchynet}
Surat Teerapittayanon, Bradley McDanel, and Hsiang-Tsung Kung.
\newblock Branchynet: Fast inference via early exiting from deep neural
  networks.
\newblock In \emph{2016 23rd International Conference on Pattern Recognition
  (ICPR)}, pp.\  2464--2469. IEEE, 2016.

\bibitem[Touvron et~al.(2021{\natexlab{a}})Touvron, Cord, Douze, Massa,
  Sablayrolles, and Jegou]{deit}
Hugo Touvron, Matthieu Cord, Matthijs Douze, Francisco Massa, Alexandre
  Sablayrolles, and Herve Jegou.
\newblock Training data-efficient image transformers and distillation through
  attention.
\newblock In \emph{International Conference on Machine Learning (ICML)},
  2021{\natexlab{a}}.

\bibitem[Touvron et~al.(2021{\natexlab{b}})Touvron, Cord, Douze, Massa,
  Sablayrolles, and J{\'e}gou]{touvron2021deit}
Hugo Touvron, Matthieu Cord, Matthijs Douze, Francisco Massa, Alexandre
  Sablayrolles, and Herv{\'e} J{\'e}gou.
\newblock Training data-efficient image transformers \& distillation through
  attention.
\newblock In \emph{International Conference on Machine Learning}, pp.\
  10347--10357. PMLR, 2021{\natexlab{b}}.

\bibitem[Vaswani et~al.(2017)Vaswani, Shazeer, Parmar, Uszkoreit, Jones, Gomez,
  Kaiser, and Polosukhin]{vaswani2017attention}
Ashish Vaswani, Noam Shazeer, Niki Parmar, Jakob Uszkoreit, Llion Jones,
  Aidan~N Gomez, {\L}ukasz Kaiser, and Illia Polosukhin.
\newblock Attention is all you need.
\newblock \emph{Advances in neural information processing systems}, 30, 2017.

\bibitem[Veit \& Belongie(2018)Veit and Belongie]{veit2018convolutional}
Andreas Veit and Serge Belongie.
\newblock Convolutional networks with adaptive inference graphs.
\newblock In \emph{Proceedings of the European Conference on Computer Vision
  (ECCV)}, pp.\  3--18, 2018.

\bibitem[Verelst \& Tuytelaars(2020)Verelst and Tuytelaars]{verelst2020dynamic}
Thomas Verelst and Tinne Tuytelaars.
\newblock Dynamic convolutions: Exploiting spatial sparsity for faster
  inference.
\newblock In \emph{Proceedings of the ieee/cvf conference on computer vision
  and pattern recognition}, pp.\  2320--2329, 2020.

\bibitem[Wang et~al.(2018)Wang, Yu, Dou, Darrell, and
  Gonzalez]{wang2018skipnet}
Xin Wang, Fisher Yu, Zi-Yi Dou, Trevor Darrell, and Joseph~E Gonzalez.
\newblock Skipnet: Learning dynamic routing in convolutional networks.
\newblock In \emph{Proceedings of the European Conference on Computer Vision
  (ECCV)}, pp.\  409--424, 2018.

\bibitem[Xie et~al.(2017)Xie, Wang, Zhang, Zhou, Xie, and
  Yuille]{xie2017adversarial}
Cihang Xie, Jianyu Wang, Zhishuai Zhang, Yuyin Zhou, Lingxi Xie, and Alan
  Yuille.
\newblock Adversarial examples for semantic segmentation and object detection.
\newblock In \emph{Proceedings of the IEEE international conference on computer
  vision}, pp.\  1369--1378, 2017.

\bibitem[Xie et~al.(2020)Xie, Zhang, Zhu, Huang, and Lin]{xie2020spatially}
Zhenda Xie, Zheng Zhang, Xizhou Zhu, Gao Huang, and Stephen Lin.
\newblock Spatially adaptive inference with stochastic feature sampling and
  interpolation.
\newblock In \emph{Computer Vision--ECCV 2020: 16th European Conference,
  Glasgow, UK, August 23--28, 2020, Proceedings, Part I 16}, pp.\  531--548.
  Springer, 2020.

\bibitem[Yang et~al.(2020)Yang, Han, Chen, Song, Dai, and
  Huang]{yang2020resolution}
Le~Yang, Yizeng Han, Xi~Chen, Shiji Song, Jifeng Dai, and Gao Huang.
\newblock Resolution adaptive networks for efficient inference.
\newblock In \emph{Proceedings of the IEEE/CVF conference on computer vision
  and pattern recognition}, pp.\  2369--2378, 2020.

\bibitem[Yang et~al.(2019)Yang, Xu, Dai, and Xiong]{yang2019dynamic}
Zerui Yang, Yuhui Xu, Wenrui Dai, and Hongkai Xiong.
\newblock Dynamic-stride-net: Deep convolutional neural network with dynamic
  stride.
\newblock In \emph{Optoelectronic Imaging and Multimedia Technology VI}, volume
  11187, pp.\  42--53. SPIE, 2019.

\bibitem[Yin et~al.(2022)Yin, Vahdat, Alvarez, Mallya, Kautz, and
  Molchanov]{yin2022vit}
Hongxu Yin, Arash Vahdat, Jose~M Alvarez, Arun Mallya, Jan Kautz, and Pavlo
  Molchanov.
\newblock A-vit: Adaptive tokens for efficient vision transformer.
\newblock In \emph{Proceedings of the IEEE/CVF Conference on Computer Vision
  and Pattern Recognition}, pp.\  10809--10818, 2022.

\bibitem[Yu et~al.(2022)Yu, Luo, Zhou, Si, Zhou, Wang, Feng, and
  Yan]{yu2022metaformer}
Weihao Yu, Mi~Luo, Pan Zhou, Chenyang Si, Yichen Zhou, Xinchao Wang, Jiashi
  Feng, and Shuicheng Yan.
\newblock Metaformer is actually what you need for vision.
\newblock In \emph{Proceedings of the IEEE/CVF Conference on Computer Vision
  and Pattern Recognition}, pp.\  10819--10829, 2022.

\bibitem[Yu et~al.(2021)Yu, Rao, Wang, Liu, Lu, and Zhou]{pointr}
Xumin Yu, Yongming Rao, Ziyi Wang, Zuyan Liu, Jiwen Lu, and Jie Zhou.
\newblock Pointr: Diverse point cloud completion with geometry-aware
  transformers.
\newblock In \emph{IEEE/CVF International Conference on Computer Vision
  (ICCV)}, 2021.

\bibitem[Yuan et~al.(2020)Yuan, Wu, Sun, Liang, Zhao, and Bi]{yuan2020s2dnas}
Zhihang Yuan, Bingzhe Wu, Guangyu Sun, Zheng Liang, Shiwan Zhao, and Weichen
  Bi.
\newblock S2dnas: Transforming static cnn model for dynamic inference via
  neural architecture search.
\newblock In \emph{Computer Vision--ECCV 2020: 16th European Conference,
  Glasgow, UK, August 23--28, 2020, Proceedings, Part II 16}, pp.\  175--192.
  Springer, 2020.

\bibitem[Zhao et~al.(2021)Zhao, Jiang, Jia, Torr, and Koltun]{pointtransformer}
Hengshuang Zhao, Li~Jiang, Jiaya Jia, Philip Torr, and Vladlen Koltun.
\newblock Point transformer.
\newblock In \emph{IEEE/CVF International Conference on Computer Vision
  (ICCV)}, 2021.

\bibitem[Zheng et~al.(2021)Zheng, Lu, Zhao, Zhu, Luo, Wang, Fu, Feng, Xiang,
  Torr, and Zhang]{setr}
Sixiao Zheng, Jiachen Lu, Hengshuang Zhao, Xiatian Zhu, Zekun Luo, Yabiao Wang,
  Yanwei Fu, Jianfeng Feng, Tao Xiang, Philip~H.S. Torr, and Li~Zhang.
\newblock Rethinking semantic segmentation from a sequence-to-sequence
  perspective with transformers.
\newblock In \emph{IEEE/CVF Conference on Computer Vision and Pattern
  Recognition (CVPR)}, 2021.

\bibitem[Zhou et~al.(2016)Zhou, Khosla, Lapedriza, Oliva, and
  Torralba]{zhou2016learning}
Bolei Zhou, Aditya Khosla, Agata Lapedriza, Aude Oliva, and Antonio Torralba.
\newblock Learning deep features for discriminative localization.
\newblock In \emph{Proceedings of the IEEE conference on computer vision and
  pattern recognition}, pp.\  2921--2929, 2016.

\bibitem[Zhou et~al.(2021)Zhou, Kang, Jin, Yang, Lian, Jiang, Hou, and
  Feng]{deepvit}
Daquan Zhou, Bingyi Kang, Xiaojie Jin, Linjie Yang, Xiaochen Lian, Zihang
  Jiang, Qibin Hou, and Jiashi Feng.
\newblock Deepvit: Towards deeper vision transformer.
\newblock \emph{arXiv preprint arXiv:2103.11886}, 2021.

\end{thebibliography}
\bibliographystyle{iclr2024_conference_arxiv}
\clearpage
\appendix
\section{Appendix}

In sections ~\ref{sec:viz_token_drop} and ~\ref{sec:viz_patch}, we provide visualizations
of our learned patches and token dropping respectively. In Sec.~\ref{sec:supp_imlp_deets},
we provide additional details on our train and test settings. We also provide the code for
our implementation with the default hyperparameters as part of the supplementary material.

\section{Patch Visualization}
\label{sec:viz_patch}

    
    
\begin{figure}[h]
    \centering
    
        \centering
        \includegraphics[width=0.6\linewidth]{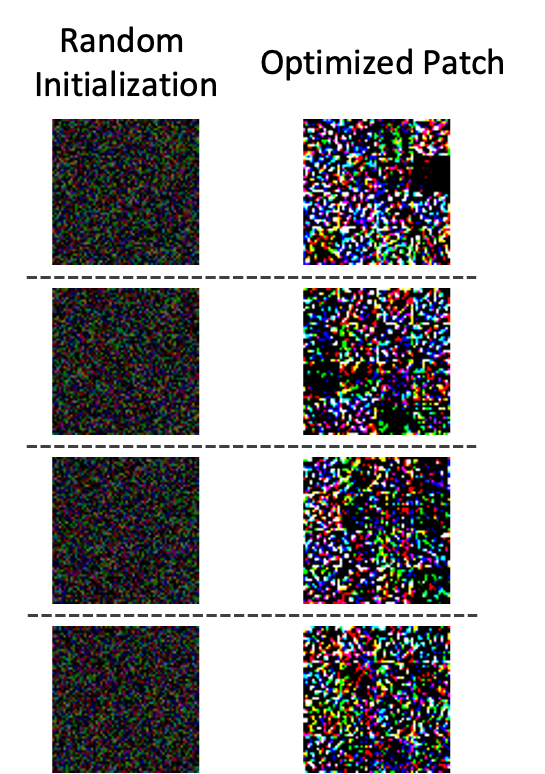}
    
    \caption{\textbf{Optimized patches With different initializations:} Here, we show the optimized 
    patches for \avit. A different initialization is used to train each of these patches. All patches
    achieve Attack Success close to $100\%$. Presence of multiple universal adversarial patches 
    highlights the vulnerability of the current efficient methods.}
    \label{fig:vis2}
\end{figure}

We optimize patches for \avit using different initializations and visualize them in Fig.~\ref{fig:vis2}.
All patches achieve Attack Success close to $100\%$. Presence of multiple universal adversarial patches 
highlights the vulnerability of the current efficient methods.

\begin{figure*}[t]
    \centering
    
        \centering
        \includegraphics[width=0.8\linewidth]{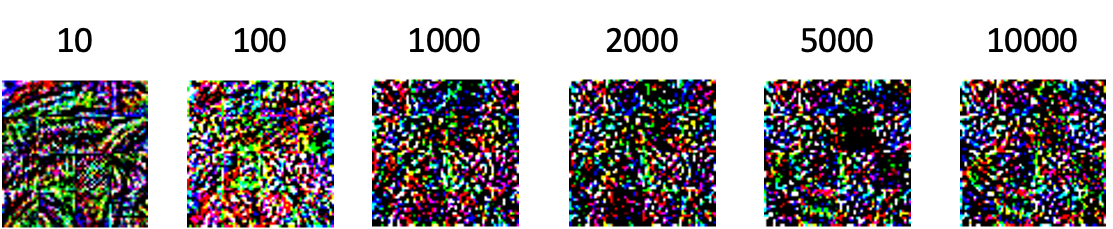}
    
    \caption{\textbf{Visualization of patch optimization:} We train our patch to attack \avit and 
    display the patch at various stages of optimization. We observe that the patch converges quickly. 
    The patch at $1000$ iterations ($0.1$ epoch) is similar to that at $10000$ iterations ($1$ epoch)
    in terms of both appearance and attack performance.}
    \label{fig:vis3}
\end{figure*}

We show the evolution of the patch as training progresses in Fig.~\ref{fig:vis3}. The patch is trained
to attack \avit approach. We observe that the patch converges quickly, requiring less than an epoch for $100\%$ 
Attack Success. The patch at $1000$ iterations ($0.1$ epoch) is similar to that at 
$10000$ iterations ($1$ epoch) in terms of both appearance and attack performance.

\section{Visualization of Token Dropping}
\label{sec:viz_token_drop}
\begin{figure*}[t]
    \centering
    
        \centering
        \includegraphics[width=1.0\linewidth]{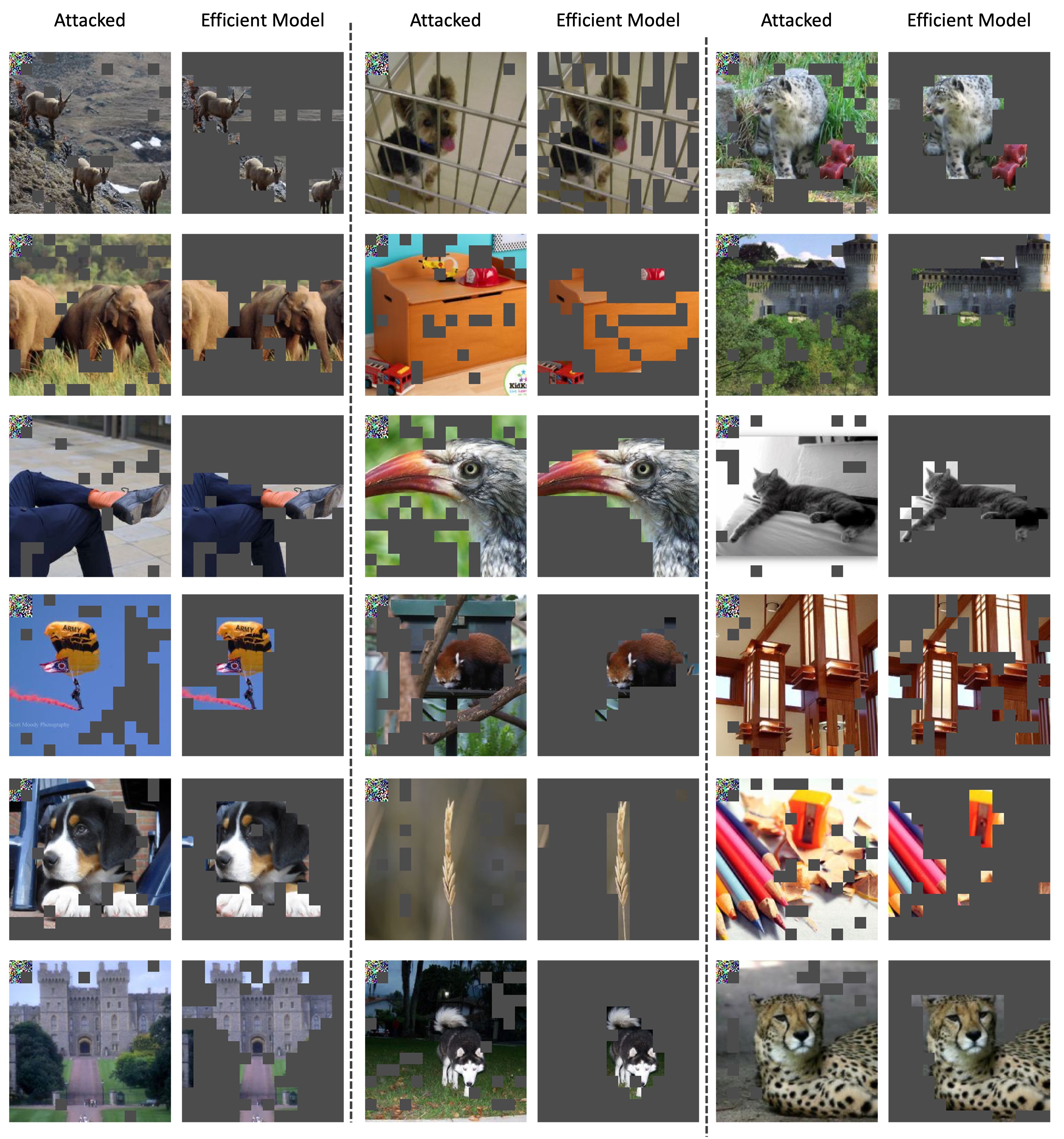}
    
    \caption{\textbf{Visualization of our Energy Attack on Vision Transformers:} Similar to Figure 2 of the main submission, we visualize the A-ViT-Small with and without our attack. We use patch size of $32$ for the attack (on the top-left corner). We show pruned tokens at layer $8$ of A-ViT-Small. Our attack can recover most of the pruned tokens, resulting in increased computation and power consumption. }
    \label{fig:vis4}
\end{figure*}

 In Fig.~\ref{fig:vis4}, we visualize dropped tokens in A-ViT-Small with and without our attack. Our attack significantly decreases the number of pruned tokens, resulting in more compute and energy consumption for the efficient transformer model.  
\section{Implementation Details}
\label{sec:supp_imlp_deets}

\noindent \textbf{ATS Details:} As in ATS~\cite{fayyaz2022adaptive}, we replace layers 3 through 9 of ViT networks with the ATS block and set the maximum limit for the number of tokens sampled to $197$ for each layer. We train the patch for 2 epochs with a learning rate of $0.4$ for ViT-Tiny and $lr=0.2$ for ViT-Base and ViT-Small. 
We use a batch size of $1024$ and different loss coefficients for each layer of ATS. For DeiT-Tiny we use $ [1.0, 0.2, 0.2, 0.2, 0.01, 0.01, 0.01] $, for DeiT-Small we use $ [ 1.0, 0.2, 0.05, 0.01, 0.005, 0.005, 0.005 ] $, and for DeiT-Base we use $ [ 2.0, 0.1, 0.02, 0.01, 0.005, 0.005, 0.005]$ The weights are vastly different at initial and final layers to account for the difference in loss magnitudes across layers. \\

\noindent \textbf{A-ViT Details:} The patches are optimized for one epoch with a learning rate of $0.2$ and a batch size of $512$ ($128\times 4$GPUs) using AdamW~\cite{loshchilov2017decoupled} optimizer. We optimize the patches for $4$ epochs for patch length $32$ and below. For CIFAR-10 experiments, the images are resized from $32\times 32$ to $256 \times 256$ and a $224 \times 224$ crop is used as the input. For the training of adversarial defense, we generate 5 patches per epoch of adversarial training and limit the number of iterations for patch generation to $500$. The learning rate for patch optimization is increased to $0.8$ for faster convergence. \\

\noindent \textbf{AdaViT Details:} We use a learning rate of $0.2$ and a batch size of $128$ with 4GPUs for patch optimization. We use AdamW~\cite{loshchilov2017decoupled} optimizer with no decay and train for 2 epochs with a patch size of 64 x 64. We train on the ImageNet-1k train dataset and evaluate it on the test set.

\end{document}